\documentclass[runningheads]{llncs}
\usepackage[T1]{fontenc}
\usepackage[utf8]{inputenc} % allow utf-8 input
\usepackage[T1]{fontenc}    % use 8-bit T1 fonts
\usepackage{hyperref}       % hyperlinks
\usepackage{url}            % simple URL typesetting
\usepackage{booktabs}       % professional-quality tables
\usepackage{amsfonts}       % blackboard math symbols
\usepackage{nicefrac}       % compact symbols for 1/2, etc.
\usepackage{microtype}      % microtypography
\usepackage{graphicx}
\usepackage{xcolor}
\usepackage{enumitem}
\usepackage{graphicx}
\usepackage{algorithm}
\usepackage{algpseudocode}
\usepackage{xcolor}
\usepackage{wasysym}
\usepackage{tikz}
\usepackage{orcidlink}
\usepackage{amssymb}

\newcommand{\algorithmicoutput}{\textbf{Output:}}
\newcommand{\OUTPUT}{\item[\algorithmicoutput]}

\spnewtheorem{observation}{Observation}{\bfseries}{\itshape}

\def\figwidth{1}
\begin{document}
\title{Prime Implicant Explanations for \\Reaction Feasibility Prediction}
\author{Klaus Weinbauer\inst{1,2}\orcidlink{0000-0002-3349-9157} \and
Tieu-Long Phan\inst{2,3}\orcidlink{0000-0002-3532-2064} \and
Peter F. Stadler\inst{2,4,5,6,7,8}\orcidlink{0000-0002-5016-5191} \and\\
Thomas G\"{a}rtner\inst{1}\orcidlink{0000-0001-5985-9213} \and
Sagar Malhotra\inst{1}\orcidlink{0000-0001-6700-4311}}

\authorrunning{K. Weinbauer et al.}

\institute{Machine Learning Research Unit, TU Wien, Vienna, Austria\and
	Bioinformatics Group, Leipzig University, Leipzig, Germany\and
	Department of Mathematics and Computer Science, University of Southern Denmark, Odense, Denmark\and
	Max Planck Institute for Mathematics in the Sciences, Leipzig, Germany\and
	Department of Theoretical Chemistry, University of Vienna, Vienna, Austria\and
	Facultad de Ciencias, Universidad National de Colombia, Bogot{\'a}, Colombia\and
	Center for non-coding RNA in Technology and Health, University of Copenhagen, Frederiksberg, Denmark\and
	Santa Fe Institute, Santa Fe, USA\and
	\email{kw@ml.tuwien.ac.at}
}
\maketitle              % typeset the header of the contribution

\begingroup
\renewcommand\thefootnote{}\renewcommand\theHfootnote{}%
\footnotetext{Presented at AIMLAI workshop at ECMLPKDD 2025.}%
\addtocounter{footnote}{-1}%
\endgroup

\begin{abstract}

  Machine learning models that predict the feasibility of chemical
  reactions have become central to automated synthesis planning. Despite
  their predictive success, these models often lack transparency and
  interpretability. We introduce a novel formulation of prime implicant
  explanations---also known as minimally sufficient reasons---tailored to
  this domain, and propose an algorithm for computing such explanations in
  small-scale reaction prediction tasks. Preliminary experiments
  demonstrate that our notion of prime implicant explanations
  conservatively captures the ground truth explanations. That is, such
  explanations often contain redundant bonds and atoms but consistently
  capture the molecular attributes that are essential for predicting
  reaction feasibility.

	\keywords{Prime Implicant Explanation \and Subgraph Explanation \and Reaction Prediction \and Synthesis Planning}
\end{abstract}

\section{Introduction}

We aim to generate formally grounded and reliable explanations for reaction
feasibility (RF) prediction tasks~\cite{Segler2017a}. Predicting
feasibility of chemical reactions is fundamental to the problem of
computer-aided synthesis planning (CASP)\cite{Corey+1967+19+38}. Although
machine learning (ML) models have shown remarkable success in this
field~\cite{Coley2018}, their use can lead to significant reduction in
transparency and interpretability~\cite{rudin2019explaining}.  To address
these challenges, initial efforts have been made to enhance the
explainability of ML based RF prediction models, for example, by combining
coarse level and fine level   representations~\cite{Hou2023} or by applying
layer-wise relevance propagation~\cite{Bach2015} in a neural network
classifier~\cite{Kim2021}. However, such explanations do not have
well-defined semantics and can often be misleading
\cite{Misleading_Explanations,rudin2019explaining}. Furthermore, to the
best of our knowledge, there is no widely accepted notion of an explanation
with formally defined semantics in the context of reaction feasibility
prediction. Here, we aim to fill this gap.

\begin{figure}[t!bh]
	\centering
	\includegraphics[width=\figwidth\textwidth]{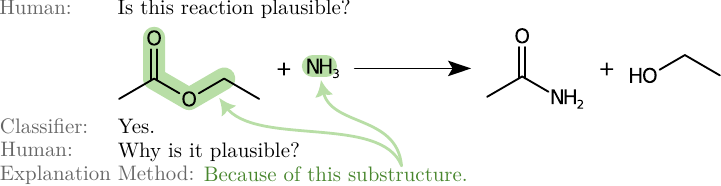}

	\caption{The proposed explanation method provides a minimally sufficient
		subgraph for the decisions made by a black-box reaction feasibility
		classifier. This subgraph explanation reflects the chemical mechanism
		in which \(\mathrm{NH}_3\) attacks the ester's electrophilic carbonyl
		carbon, displacing the alkoxide and forming the more
		resonance-stabilized amide.}
	\label{fig:pi_example}
\end{figure}

We leverage and expand a formal notion of minimally sufficient explanations
known as \emph{prime implicant (PI) explanations}. PI explanations are a
formally grounded and widely accepted notion of explanations for
propositional logic classifiers \cite{Darwiche2020}. Recent results have
devised a similar notion of explanations for graph classification tasks
\cite{Azzolin2025}. However, formal notion of explanations for applications
similar to RF prediction are relatively underexplored. We first show that
conventional notions of PI explanations \cite{Azzolin2025} are semantically
misaligned and computationally expensive for RF prediction tasks. We then
introduce \emph{PI reaction explanations} for explaining RF prediction
tasks. Although computing PI reaction explanations is intractable in
general, we show that for small instances they can be identified
efficiently. Our method uses the Imaginary Transition State graphs (ITS),
i.e., node and edge labeled graphs that provide a succinct representation
of a chemical reaction. We show that explaining RF predictions reduces to
finding minimally sufficient rooted connected subgraphs of an ITS. To solve
this search problem, we introduce \emph{extension DAGs} a directed acyclic
graph (DAG) that encodes the partial order induced by rooted
subgraphs of an ITS. We perform preliminary empirical analysis using expert
ratings on 179 RF prediction explanations from a graph neural network based
classifier. Our analysis shows that PI reaction explanations consistently
capture the ground truth explanation as a subgraph but often include
multiple non-essential nodes and edges. We conclude with a brief discussion
on the potential reasons behind the observed redundancies, open challenges,
and future directions.

\section{Preliminaries} 

This section introduces the foundational concepts
and terminology needed to formalize the problem and present the proposed
solution. We begin by establishing basic graph-theoretic notions, give an
overview of prime implicant (PI) explanations, and describe the chemical
reaction representation used in graph classification.

Chemical notation specifies molecules as labeled graphs $G$, with vertices
denoting atoms and edges denoting chemical bonds between them. We
write $V(G)$ and $E(G)$ for the node and edge set of the graph $G$, and
assume that $G$ belongs to a graph class of interest $\mathcal{G}$. We
write edges as $uv\in E(G)$, and in undirected graphs $uv = vu$. A subgraph
$H$ of $G$ is denoted $H\subseteq G$, and the node-induced subgraph of $G$
on the node set $V(H)$ is written as $G[V(H)]=H$. Similarly, the subgraph
of $G$ consisting of the edges of $H$ is denoted by $G[E(H)]$.

\paragraph{Prime implicant explanation} 

A prime implicant (PI) explanation of a classification is a subset of
features from the instance that is minimally sufficient for the
prediction~\cite{Shih2018}. For decision problems,  a PI explanation is a
minimal set of features implying the decision.  PI explanations are the
analogue of prime implicants in propositional logic. For example, consider
the function $f(x, y, z) = (x \wedge y) \vee \bar{z}$ as a binary
classifier for the features $x, y, z$. The PI explanations for $f$ are
$x\wedge y$ and $\bar{z}$. PI explanations give an intuitive description of
what the classifier considers sufficient for the respective prediction. PI
explanations are not unique in general.

\paragraph{Imaginary Transition State graph} 

The Imaginary Transition State (ITS)~\cite{Fujita1986,Wilcox1986} graph is
a graph-theoretical representation of the structural changes that
molecules, encoded as annotated graphs, undergo during a chemical reaction.
It is the superposition of the molecular graphs of reactants and products,
where edge labels encode the changes in bond order between the atoms. Thus,
the edge labels in the ITS graph are tuples, with the first element
encoding the bond order in the reactant graph and the second element
encoding the bond order in the product graph. Edges with unequal label
values encode a transformation. In the chemical domain, this corresponds to
the formation or breaking of valence bonds. We use the ITS graph as defined
in \cite{GonzalezLaffitte2024}, see Appendix~\ref{def:its_graph} for the
formal definition.  Every ITS graph contains a reaction center, defined as
the subgraph consisting of the changing bonds. It represents the structural
transformation that occurs during the chemical reaction.

\begin{definition}[Reaction Center]
	The reaction center $\Gamma$ of an ITS graph
	$G$ with edge labels $(b^e_R, b^e_P)$ for edge $e \in E(G)$ is defined by
	$\Gamma := G[\{e | b^e_R \neq b^e_P\}]$.
\end{definition}

\begin{figure}[t!bh]
	\centering
	\includegraphics[width=\figwidth\textwidth]{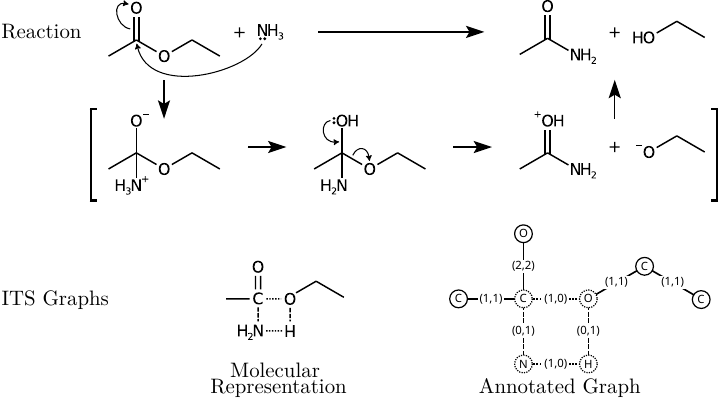}

	\caption{Two ITS graph visualizations for the reaction in the first row.
		We'll mostly use the more concise molecular representation on the left
		but the ITS graph is actually a node and edge labeled graph as shown on
		the right. The reaction center is the subgraph of changing bonds and is
		visualized using dotted and dashed lines. Dotted lines indicate breaking
		bonds, while dashed lines represent newly formed bonds.}

	\label{fig:example_its}
\end{figure}

Fig.~\ref{fig:example_its} depicts the ITS graph of a chemical reaction.
Besides the annotated ITS graph, the figure also shows a concise,
molecular-graph-like representation that we will be using. The reaction
center is marked by dotted and dashed lines. The ITS graph is a convenient
representation of chemical reactions because all for the reaction relevant
structures are adjacent to the reaction center, and hence the ITS graph is
connected~\cite{GonzalezLaffitte2024}.

\paragraph{Reaction feasibility prediction}

\begin{figure}[t!bh]
	\centering
	\includegraphics[width=\figwidth\textwidth]{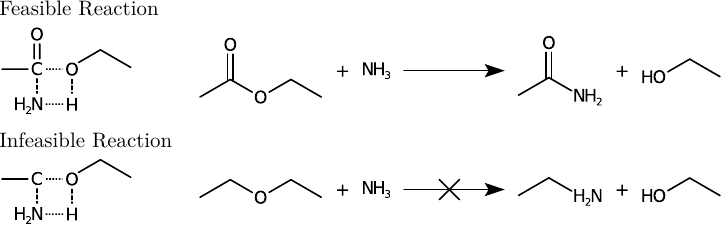}

	\caption{Example of two ITS graphs with isomorphic reaction center. The
		ITS graph representing the first reaction is feasible whereas the second
		ITS graph represents an infeasible reaction. The reaction center alone is
		insufficient to decide the feasibility of a reaction. Some context
		is needed.}

	\label{fig:its_plausibility_example}
\end{figure}

Reaction feasibility (RF) prediction is a binary graph classification
problem on ITS graphs, where the prediction target is the chemical
feasibility of a given ITS graph. These predictions could, for example,
improve template-based synthesis planning by ruling out chemically
infeasible combinatorial solutions. Fig.~\ref{fig:its_plausibility_example}
depicts two reactions---one feasible and the other infeasible. Although the
reaction centers in both cases are isomorphic, the feasibility of the
reaction depends on some additional surrounding context. To evaluate these
classifier decisions, we introduce and analyze subgraph PI explanations.
These explanations allow domain experts to assess whether the classifier
has learned chemical domain knowledge.

\section{PI Explanations on Graphs}

We begin by adapting the definition of a PI explanation by Shih et
al.~\cite[Def.5]{Shih2018} to the setting of simple, undirected graphs,
following an approach similar to \cite{Azzolin2025}. In a second step, we
further adapt this definition to the domain-specific setting of reaction
feasibility prediction.

\pagebreak
\begin{definition}[Subgraph PI explanation~\cite{Azzolin2025}]
	\label{def:pi_graph_explanation} Let $h:\mathcal{G}\to \{0, 1\}$ be the
	binary classification function and $G\in \mathcal{G}$ the graph instance.
	A PI explanation is a graph $Z$ such that
	\begin{enumerate}[label=(\alph*),ref={(\alph*)}]
		\item $Z\subseteq G$,
		\item \label{defenum:extensions} $h(Z')=h(G)$ for all $Z \subseteq Z' \subseteq G$,
		\item and no proper subgraph $Z''\subset Z$ satisfies (a) and (b).
	\end{enumerate}
\end{definition}

\begin{remark} A key difference of subgraph PI explanations to PI
	explanations in general is in point~\ref{defenum:extensions}. In the
	general case, all possible extensions from the domain are allowed $Z
		\subseteq Z'$. In contrast, subgraph PI explanations only permit
	extensions within the instance $Z \subseteq Z' \subseteq G$. This
	constraint is necessary to reduce the complexity of finding PI
	explanations, as the general case is highly intractable.

\end{remark}

A subgraph PI explanation for a binary graph classification is a subgraph
such that all of its supergraphs within the instance are assigned the same
label. To find all PI explanations, one can, in principle, exhaustively
query the classifier on all possible subgraphs and select those that
satisfy Definition~\ref{def:pi_graph_explanation}. This brute-force
approach is intractable in practice because the number of subgraphs
grows exponentially with graph size. However, due to the nature of the
application domain, the space of admissible graphs $\mathcal{G}$ can be
constrained. Although these constraints do not improve the worst-case
complexity, they substantially reduce the number of subgraphs, making the
problem solvable for small instances, as demonstrated in the empirical
complexity analysis in Appendix~\ref{asec:complexity}.

\subsection{PI Reaction Explanation}

We now adapt the previously introduced subgraph PI explanations to a
domain-specific setting for obtaining explanations from RF predictions,
guided by two key domain observations. Within the context of chemical
reactions encoded as ITS graphs, two graph class restrictions can be
derived from prior work on their structural
properties~\cite{GonzalezLaffitte2024}. Firstly, in an ITS graph, all
compound graphs undergoing bond changes are connected to the reaction
center~\cite{GonzalezLaffitte2024}. Environmental effects that are
transparent to structural changes---such as solvents---are inherently
absent from this representation. Likewise, non-valence interactions such as
hydrogen bonds are not captured in ITS graphs. Consequently, extensions
must be connected graphs, and thus, all disconnected subgraphs can be
discarded. Secondly, each ITS graph contains a reaction center---the
subgraph that encodes all the bond changes between reactants and
products~\cite{GonzalezLaffitte2024}. This component is essential for the
graph to represent a reaction mechanism. Without it, the ITS graph does not
encode any chemical transformation. Hence, PI reaction explanations
necessarily are connected subgraphs rooted at the reaction center.

\begin{definition}[PI reaction explanation]
	\label{def:pi_reaction_explanation}
	Let $h:\mathcal{G}\to \{0, 1\}$ be a reaction feasibility classifier, and
	$G\in \mathcal{G}$ be an instance from the class of connected ITS
	graphs with $\Gamma$ denoting its reaction center. A PI reaction
	explanation is a graph $Z$ such that
	\begin{enumerate}[label=(\alph*)]
		\item $\Gamma \subseteq Z \subseteq G$,
		\item \label{defenum:rxn_pi_extensions} $h(Z') = h(G)$ for all $Z \subseteq Z' \subseteq G$
		\item $Z$ is connected,
		\item and no proper subgraph $Z'' \subset Z$ satisfies (a) to (c).
	\end{enumerate}
\end{definition}

The connectivity constraint, together with the necessary root, reduces the
problem from enumerating all subgraphs to enumerating only rooted connected
subgraphs.

\subsection{Rooted and Connected Subgraph Extensions}

Computing PI reaction explanations for an ITS graph involves classifying
its rooted connected subgraphs. To reduce the number of classifier
decisions, we utilize a directed acyclic graph (DAG) defined by a partial
ordering. The nodes in this extension DAG are the set of rooted connected
subgraphs, and the edges correspond to the subgraph-supergraph
relationships. This structure captures how extensions can be incrementally
constructed.

\begin{definition}[Extension DAG] Let $G$ be an instance, and let $S$ be
	the set of connected subgraphs of $G$ that contain the root as subgraph.
	For $\vartheta, \upsilon \in S$ we say that $\vartheta \preceq \upsilon$
	iff $\vartheta$ is a subgraph of $\upsilon$. Then the extension DAG
	$\mathcal{D}$ represents the partial order (lattice).

\end{definition}

That is, $\mathcal{D}$ is a DAG with nodes $V(\mathcal{D})=S$ and
edges $E(\mathcal{D}) = \{ (\vartheta, \upsilon) \in (S \times S) |
	\vartheta \prec \upsilon$ and there is no $z \in S$ such that $\vartheta
	\prec z \prec \upsilon \}$.  In other words, $\mathcal{D}$ is the Hasse
diagram for the subgraph relation on the possible PI explanations.

\begin{observation}
	An extension DAG for the instance $G$ contains a unique sink node,
	representing the root, and a unique source node, which corresponds to the
	entire instance $G$.
\end{observation}

\begin{figure}[t!bh]
	\centering
	\includegraphics[width=\figwidth\textwidth]{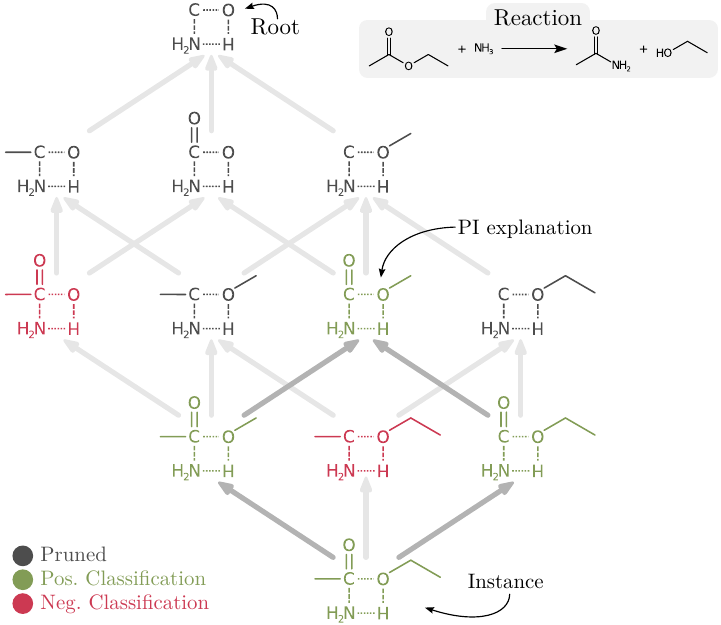}

	\caption{Example for computing PI reaction explanations. First,
		Algorithm~\ref{alg:extension_DAG} constructs the extension DAG. It starts
		with the reaction center (the root) and iteratively constructs the
		connected extensions up to the instance. Next,
		Algorithm~\ref{alg:compute_pi} computes the PI explanations by evaluating
		the extensions starting from the instance. Once the algorithm encounters
		a negative prediction, it prunes all paths upwards. The PI explanations
		are the nodes without any successors in the remaining DAG. There is only
		one in the example. }

	\label{fig:extension_DAG}
\end{figure}

\paragraph{Extension DAG construction}

The algorithm for constructing the extension DAG of $G$ is adapted from the
linear-delay enumeration algorithm for connected node-induced subgraphs
proposed by Alokshiya et al.~\cite{Alokshiya2019}, which in turn is based
on the reverse search algorithm introduced by
Avis~and~Fukuda~\cite{Avis1996}. Our adaptation modifies this approach such
that it generates a directed acyclic graph of rooted subgraphs.  The
algorithmic solution requires the root to be a single node. Without loss of
generality for the construction of the DAG we can assume that the root
subgraph is merged into a single root node by contracting its edges.
Algorithm~\ref{alg:extension_DAG} in the appendix provides a pseudocode
description of the proposed algorithm.

\subsection{Computing PI Reaction Explanations}

To address the core objective of finding PI explanations for reaction
feasibility prediction on the ITS graph $G$, we traverse its full extension
DAG $\mathcal{D}$ and classify the subgraphs. Since reaction
feasibility is a non-monotonic property~\cite{Alon2005} of the ITS graph
(see Appendix~\ref{asec:rfasitsporp}), the use of more efficient search
algorithms is limited. During this traversal, invalid paths are pruned
based on the obtained decision, yielding a final DAG in which the DAG-nodes
without descendants correspond to the PI reaction explanations of $G$.
Since the prediction on $G$ is the decision being explained, $G$ qualifies
as an explanation, although it is typically not a PI explanation. Moreover,
all subgraphs along the path to a PI explanation must preserve the
classification label. Therefore, when an extension causes a change in
prediction, we prune all of its subgraphs from $\mathcal{D}$, as any
further subgraph would violate point~\ref{defenum:rxn_pi_extensions} of
Definition~\ref{def:pi_reaction_explanation}. The algorithm terminates once
no further nodes remain to be evaluated. Algorithm~\ref{alg:compute_pi}
provides a pseudocode description for computing PI explanations based on an
extension DAG. 

\begin{algorithm}
	\caption{Algorithm for computing PI explanations
		based on an extension DAG.}
	\label{alg:compute_pi}
	\begin{algorithmic}[1]
		\Require An extension DAG $\mathcal{D}$ for the instance $G$ and a decision function $f$
		\OUTPUT The set of PI explanations of $G$

		\State $U \gets$ source nodes of $\mathcal{D}$
		\While{$|U| > 0$}
		\State $U' \gets \{\}$
		\For{$\vartheta \in U \cap V(\mathcal{D})$}
		\If{$f(G[\vartheta]) = 1$}
		\State $U' \gets U' \cup \{ \upsilon \in V(\mathcal{D}) \mid (\vartheta, \upsilon) \in E(\mathcal{D}) \}$ \Comment{add outneighbors of $\vartheta$}
		\Else
		\State remove $\vartheta$ and all descendants of $\vartheta$ from $\mathcal{D}$
		\EndIf
		\EndFor
		\State $U \gets U'$
		\EndWhile
		\State \Return $\{ G[\vartheta] | \vartheta \in$ sink nodes of $\mathcal{D} \}$ \Comment{nodes in $\mathcal{D}$ are node sets of $G$}
	\end{algorithmic}
\end{algorithm}

Fig.~\ref{fig:extension_DAG} shows the PI reaction explanation search on
the extension DAG. Evaluation starts at the full instance with a positive
prediction. In the next iteration, all outneighbor extensions from the
previous round are tested. In the example, there are two positive
predictions and one negative prediction. For the negative prediction the
algorithm prunes all its descendants, i.e., all paths upwards. After
another iteration, the algorithm terminates. The nodes (only one in the
example) without descendants in the remaining DAG are the PI reaction
explanations.

To find PI reaction explanations for ITS graph classifications, we consider
connected and rooted edge-induced subgraphs, which best capture the
chemical intuition behind molecular
interactions~\cite{Benkoe2003,Flamm2023}. In molecular orbital theory,
atoms can alter the properties of a molecule by sharing electrons through
overlapping orbitals---the valence bonds. The presence or absence of a
valence bond can significantly influence a molecule's electronic structure
and, consequently, its reactivity. In the ITS graph representation, valence
bonds and bond changes are modeled as edges. Thus, the presence or absence
of a single edge can render a reaction mechanism feasible or infeasible, as
illustrated in Fig.~\ref{fig:its_plausibility_example}. An additional edge
can allow an otherwise disconnected atom to contribute its electrons and
enable the reaction. Therefore, it is more natural and chemically
meaningful to base our explanations on edge sets rather than node sets.
Although enumerating edge-induced subgraphs differs technically from
enumerating node-induced subgraphs, this distinction can be bridged by
employing the line graph $L(G)$ of a graph $G$. Each edge-induced subgraph
in $G$ corresponds to a node-induced subgraph in $L(G)$, and vice versa.
Hence, enumerating edge-induced subgraphs in $G$ is equivalent to
enumerating node-induced subgraphs in $L(G)$. The extension DAG used to
compute PI reaction explanations for an ITS graph $G$ is constructed using
Algorithm~\ref{alg:extension_DAG} on $L(G)$ after contracting the reaction
center edges.

\section{Evaluation}

To evaluate our approach, we construct a small dataset for reaction
feasibility prediction using the publicly available USPTO reaction
dataset~\cite{Jin2017,Lowe2012}. We train a Graph Isomorphism
Network~(GIN)~\cite{Xu2019} for graph classification and derive PI reaction
explanations for its decisions. Due to the lack of an established
	benchmark for evaluating PI explanations in this context, we perform a
	qualitative analysis, comparing the results against expert judgment of the
	underlying structural reasons.

\begin{figure}[t!bh]
	\centering
	\includegraphics[width=\figwidth\textwidth]{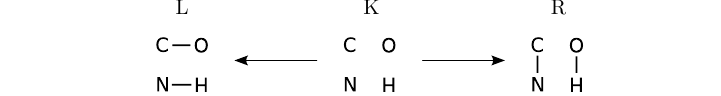}

	\caption{The double pushout rule that is used to construct the dataset.
		Single bond between Carbon and Oxygen is broken and a new single bond
		between Nitrogen and Carbon is formed. Hydrogen is shown for
		completeness but is not represented as explicit node in the dataset.}

	\label{fig:reaction_rule}
\end{figure}

\paragraph{Dataset} The dataset is generated from the USPTO-MIT
\cite{Jin2017} dataset by applying a reaction template on the reactant
molecules from the dataset and selecting a balanced subset of positive and
negative instances from the obtained reaction candidates. A reaction
candidate is considered positive if the candidate product matches the
expected product from the data, and negative otherwise. As complexity of
the proposed method is exponential in the size of the graph we consider ITS
graphs of at most 25 nodes. For the moment, we restrict ourselves to one
specific reaction mechanism, namely the substitution of Oxygen by Nitrogen
at a Carbon atom as depicted in Fig.~\ref{fig:reaction_rule}.  The
USPTO-MIT contains 3809 reactions with this reaction center, which serve as
positive examples. Negative samples are generated by undersampling an
equal-sized set from the infeasible candidates. The final dataset therefore
contains 7618 ($50\%$ positive and $50\%$ negative) instances.

\begin{figure}[t!bh]
	\centering
	\includegraphics[width=\figwidth\textwidth]{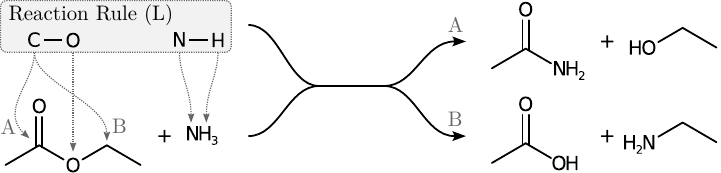}

	\caption{Reaction rule application on reactant molecule. A rule can have
		multiple matches in the reactant graph and hence can yield multiple
		reaction candidates. The correct product is A. The ITS graph of
		reaction B is infeasible.}
	\label{fig:rule_application}
\end{figure}

Applying a reaction rule to a reactant molecular graph is a combinatorial
process that can produce multiple possible reaction candidates, as
illustrated in Fig.~\ref{fig:rule_application}. These candidates are
generated by identifying subgraph isomorphisms between the left-hand side
graph $L$ of the rule and the reactant graph. The ITS graphs can directly
be constructed by this alignment and the right-hand side graph $R$ of the rule.

\subsection{Experiment}

For the experimental evaluation the Graph Isomorphism
Network (GIN)~\cite{Xu2019} is chosen as classifier for its strong
performance on graph-based learning tasks and, crucially, its fast
inference. The latter is particularly important because computing PI
explanations requires a large number of predictions. We perform a
stratified dataset split of 60\% for training, 20\% for validation, and
20\% for testing. Model selection is guided by hyperparameter optimization
on the training and validation sets, using the tree-structured Parzen
estimator~(TPE)~algorithm~\cite{Bergstra2011} as implemented in the
optimization framework Optuna~\cite{Akiba2019}. Based on preliminary
experiments we fix the number of training epochs to $100$ and the batch
size to $1024$. We then optimize the following hyperparameters in the
respective ranges over $100$ iterations: number of layers $\{2, 3, 4, 5\}$,
layer size $\{8, 16, 32, 64, 128, 256\}$, dropout $[0, 0.8]$, readout
function $\{ max, sum \}$, and learning rate $[0.0001,0.005]$. The best
model configuration includes $5$ layers of size $32$ with a dropout rate of
$0.04$, a $max$ readout function, and a learning rate of $0.003$. The final
model configuration is trained 10 times on the combined training and
validation sets, and its performance is evaluated on the held-out test set.
The test performance is summarized in Table~\ref{tab:test_results}. The
table shows the mean and standard deviation of accuracy, precision, recall,
F1 score and the area under the receiver operator curve (AUROC) in percent.
The values are averaged over 10 runs.

\setlength{\tabcolsep}{10pt}
\begin{table}[htb]
	\caption{Test performance of GIN in \% averaged over 10 runs.}
	\label{tab:test_results}
	\centering
	\begin{tabular}{ccccc}
		\toprule
		Accuracy      & Precision     & Recall        & F1 score      & AUROC         \\
		\midrule
		$86.1\pm 2.1$ & $85.6\pm 3.4$ & $87.0\pm 2.1$ & $86.3\pm 1.8$ & $93.2\pm 1.4$ \\
		\bottomrule
	\end{tabular}
\end{table}

With the trained model in place we compute the PI reaction explanations for
each true positive instance in the test set. For this, we construct the
extension DAG based on Algorithm~\ref{alg:extension_DAG} and compute the
explanations according to Algorithm~\ref{alg:compute_pi}. To improve
inference time, we can utilize GPU-based batch processing of extensions.
Instead of querying the classifier with individual instances we can
assemble a batch from the loop in lines 4-10 of
Algorithm~\ref{alg:compute_pi}.

\subsection{Analysis of Results}

The qualitative analysis---based on 179 manually identified expected
explanations from the 653 true positive predictions---revealed that the PI
reaction explanations significantly diverge from expert mechanistic
interpretations. A chemically sound reaction explanation must capture
changes in all relevant atoms and bonds throughout the mechanistic
sequence, including necessary neighboring atoms. For instance, any valid
explanation of the nucleophilic addition-elimination reaction in
Fig.~\ref{fig:example_its} must include the carbonyl double bond, essential
for both forming the tetrahedral intermediate and cleaving the C-O bond
during elimination. Since the classifier's PI explanations often lack
precision in reflecting the true underlying chemical mechanisms, we adopt a
1-to-6 rating scale instead of a binary correct/incorrect label. This
allows us to capture varying degrees of alignment between the model's
reasoning and established chemical knowledge. A rating of 1 indicates the
best match, where the obtained explanation exactly matches the expected
explanation. A rating of 6 indicates the worst match, where the expected
explanation is not even a subgraph of the obtained explanation. The
categories are defined as follows.

\begin{enumerate}

	\item \textbf{Perfect explanation:} The explanation is isomorphic to the
	      expected explanation.

	\item \textbf{Good explanation}: The explanation is a supergraph of the
	      expected explanation and has at most three additional carbon atoms.

	\item \textbf{Acceptable explanation}: The explanation is a supergraph of
	      the expected explanation and has at most five additional carbon atoms
	      and at most one additional Nitrogen or Oxygen atom.

	\item \textbf{Mediocre explanation}: The explanation is a supergraph of
	      the expected explanation and has at most eight additional carbon atoms
	      and at most two additional Nitrogen or Oxygen atoms.

	\item \textbf{Bad explanation}: The explanation is a supergraph of the
	      expected explanation but there are too many additional atoms.

	\item \textbf{Not an explanation}: The expected explanation is not a
	      subgraph of the explanation.

\end{enumerate}

\begin{figure}[t!bh]
	\centering
	\includegraphics[width=\figwidth\textwidth]{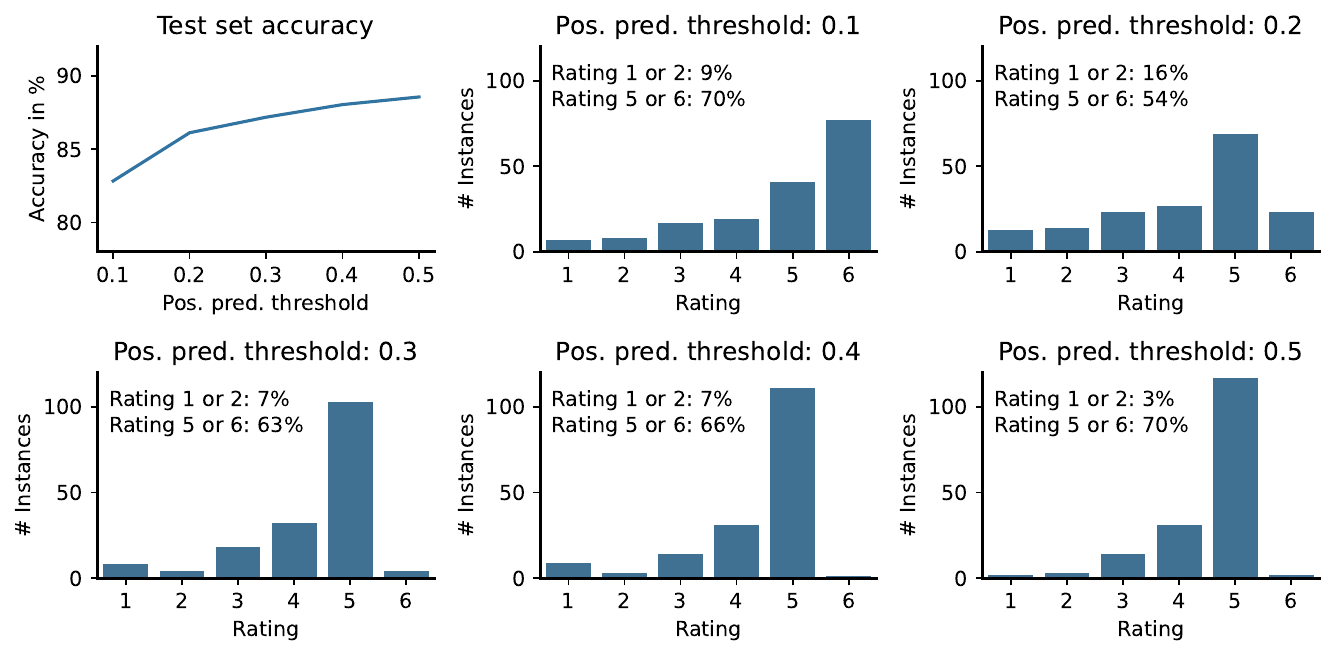}

	\caption{Results of the empirical evaluation. The top-left plot shows the
		accuracy trade-off for decreasing positive prediction thresholds. The
		remaining plots depict the rating distribution for thresholds between
		$0.1$ and $0.5$. A positive prediction threshold of $0.2$ led to the
		explanations that best matched the expected explanations, with $16\%$
		achieving a good rating of $1$ or $2$, and $54\%$ achieving a bad rating
		of $5$ or $6$. }

	\label{fig:result_plot}
\end{figure}

The experiment shows that the reaction explanations are sensitive to
uncertainty in the classifier's predictions. In the default setting---where
the threshold for a positive prediction is 0.5 and the classifier outputs a
probability in the range $[0,1]$ indicating the feasibility of a
reaction---only $3\%$ of the obtained explanations receive a rating of $1$
or $2$, while $70\%$ receive a rating of $5$ or $6$. To account for
uncertain negative classifications, we gradually lower the positive
classification threshold from $0.5$ to $0.1$ in steps of $0.1$. The
accuracy decreases from $88.6\%$ to $82.8\%$, while the distribution of
explanation ratings changes significantly. As the threshold decreases, the
proportion of good ratings increases and that of bad ratings decreases. The
closest alignment with the expected explanations is observed at a threshold
of $0.2$. A threshold of $0.1$ causes a significant increase in
explanations rated $6$. Many of these explanations tend to be small, often
consisting only of the reaction center, which suggests that the bias toward
positive predictions is too strong. A summary of these results is shown in
Fig.~\ref{fig:result_plot}.

\begin{figure}[t!bh]
	\centering
	\includegraphics[width=\figwidth\textwidth]{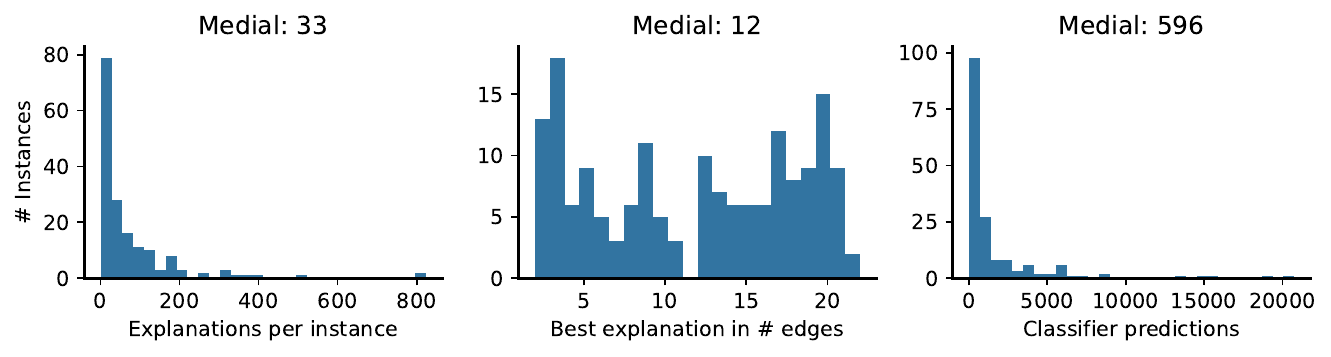}

	\caption{Distribution of explanation properties for a positive predictive
		threshold of $0.2$. The left plot shows the number of explanations per
		instance. The center plot depicts the size of the top-rated explanation,
		measured by the number of edges. The right plot shows the distribution of
		classifier predictions required to compute PI reaction explanations for
		an instance.}

	\label{fig:explanation_stats_plot}
\end{figure}

Fig.~\ref{fig:explanation_stats_plot} depicts distributions of PI reaction
explanations for the positive prediction threshold of $0.2$. The first plot
shows the distribution of the number of explanations per instance. The
median number of explanations per instance is $33$, with a maximum of
$823$. The center plot shows the distribution of the top-rated explanation
sizes in terms of the number of edges. Since hydrogen are not
explicitly represented in the dataset, the smallest possible explanation
size is $2$, i.e., the number of edges in the reaction center excluding the
hydrogen atom. The right plot shows the number of classifier decisions
required to compute PI reaction explanations. This highlights the need for
fast inference, as the number of predictions can be high for individual
instances.

Given the absence of chemically insightful PI reaction explanations, we
hypothesize three factors contributing to the obtained results.
\textbf{Inadequacy of the graph class}: The defined graph class may not
ideally represent the chemical space. While considering connected rooted
subgraphs as possible extensions is a reasonable constraint compared to the
general case of subgraphs, more chemically-aware rules might be needed to
filter out synthetically infeasible structures. It is unknown how many
genuinely infeasible extensions are generated under this model. There might
occur truly negative extensions that, if classified accordingly, would
legitimately prune the extension DAG. The expected explanations are
typically small.  This implies that, even with a perfect classifier,
successfully identifying the expected PI explanations requires a flawless
extension model---one that never generates truly negative extensions in
which the expected explanation appears as a subgraph. \textbf{Ambiguity in
ground-truth}:  The ground truth for RF prediction is inherently vague,
making it difficult to evaluate reliably. The chosen 1-to-6 evaluation
scale reflects the high degree of divergence from expected outcomes.
Determining reaction feasibility is a complex task. Manual curation
requires chemical expertise, which inherently introduces subjectivity and
leads to significant inter-annotator variability~\cite{Corbett2018}.
\textbf{Classifier limitations}: The GIN, combined with the tested readout
functions, may not be well-suited for this evaluation task. The relatively
small training set, combined with sensitivity to out-of-distribution
extensions in respect to training samples, could contribute to arbitrary
explanations. Since the number of truly infeasible ITS graphs occurring
within the extensions is unknown, measures to mitigate the effect of
out-of-distribution samples could help improve the robustness of the
predictions.

\section{Outlook}

We summarize the main challenges identified in this early study on PI
explanations for RF prediction and conclude with their
current capabilities, limitations and directions for future research.

\paragraph{Open challenges} In addition to the domain-specific difficulties
encountered during evaluation, there are also fundamental challenges
inherent to PI explanations for graph classification. \textbf{Computational
intractability}: The number of subgraph extensions grows exponentially with
the size of the instance. This limits explanations to small instances and
classifiers with fast inference times. It will be of interest to explore
whether the exhaustive enumeration can be replaced by a sampling-based
approach focussing on subgraphs whose edge-deletions preserve the
classification. The lattice structure of $\mathcal{D}$ may be helpful to
guide heuristic approximations by providing clearly defined largest common
subgraphs and smallest common supergraphs for alternative explanations.
\textbf{Lack of benchmarks}: Currently, there is no standard way for
accessing the quality of PI explanations in graph classification tasks. In
addition to limited comparability, it can be challenging to define
meaningful expected explanations---particularly in applied domains such as
cheminformatics. For instance, in the task of RF prediction, it may not
always be possible to clearly identify a substructure that is fully
accountable for a mechanism. Moreover, atoms that are not strictly
necessary for the reaction may still influence important properties such as
rate or yield, making it difficult to draw a clear boundary between
relevant and irrelevant atoms.

\paragraph{Conclusion} We introduced subgraph PI explanations for graph
classification and applied this method to explain RF predictions.
	Although limitations and open challenges remain (as discussed), PI
	explanations provide valuable insights into classifier decisions. In our
	application, these explanations correspond to the minimally sufficient
	reaction context—an inferred extension beyond the explicitly defined
	reaction center in the ITS graph. While our initial experiment shows that
	PI explanations capture important aspects of chemical intuition, they do
	not align well with expert-defined mechanistic explanations. This
	misalignment is amplified by suboptimal classifier performance and inherent
	sensitivity to noisy decisions, leading to imprecise comparisons. However,
	this comparison assumes that PI explanations inherently represent the same
	thing as the expert annotation, which may not hold true. Consequently, PI
	reaction explanations could still provide valuable insight into the model's
	reasoning. We hope these early results stimulate further research into PI
	explanations for graph data and contribute to a deeper understanding of
	chemical reasoning within reaction prediction models.

\begin{credits} \subsubsection{\ackname}

	Part of this work was funded by the European Unions Horizon Europe Doctoral
	Network program under the Marie-Sk{\l}odowska-Curie grant agreement
	No~101072930 (TACsy---Training Alliance for Computational systems chemistry).
	TG has been partially supported by the Vienna Science and Technology Fund
	(WWTF), project ICT22-059 (StruDL). PFS acknowledges support by the
	German Federal Ministry of Education and Research BMBF through DAAD project
	57616814 (SECAI, School of Embedded Composite AI). DeepL was used for
	grammar and spellchecking. ChatGPT was used to improve sentence structure
	for clarity and readability.

	\subsubsection{\discintname} Views and opinions expressed are however
	those of the author(s) only and do not necessarily reflect those of the
	European Union. Neither the European Union nor the granting authority can
	be held responsible for them.

\end{credits}

\bibliographystyle{splncs04}
\bibliography{bibliography.bib}

\clearpage
\appendix

\section{Appendix}

\subsection{Definitions}

The definition of the ITS graph used in this work follows that of Laffitte
et al.~\cite{GonzalezLaffitte2024}. The reactant and product graphs
$G$ and $H$ are disjoint unions of the relevant molecular graphs. Note
that the same molecule, i.e., connected subgraphs, may appear more than
once as a connected component in $G$ and $H$, as in the case of the
oxyhydrogen reaction 2H$_2$ + O$_2$ $\to$ 2H$_2$O. In addition to the
graph $G$ and $H$, a complete specification of a reaction also requires
the atom-atom map $\alpha:V(G)\to V(H)$ that identifies each product atom
with a corresponding reactant atom.

\begin{definition}[ITS graph~\cite{GonzalezLaffitte2024}]
	\label{def:its_graph}
	Let $\alpha : V(G) \to V(H)$ be a label
	preserving bijective function (the atom-atom map) for the reactant graph
	$G$ and the product graph $H$. Then, the Imaginary Transition State (ITS)
	$\Upsilon := \Upsilon(G, H, \alpha)$ is the graph defined by the
	following:

	\begin{enumerate}[label=(\alph*)]
		\item There is a bijection $\eta:V(\Upsilon) \to V(G)$,
		\item For $x,y\in V(\Upsilon)$ we have $xy\in E(\Upsilon)$ iff $\eta(x)\eta(y)\in E(G)$ \\
		      or $\alpha(\eta(x))\alpha(\eta(y))\in E(H)$,

		\item $xy\in E(\Upsilon)$ is labeled by the pair with labeling function
		      $b$
		      \begin{enumerate}
			      \item[(i)] $\bigg(b_G\big(\eta(x)\eta(y)\big), b_H\Big(\alpha\big(\eta(x)\big)\alpha\big(\eta(y)\big)\Big)\bigg)$
			            if $\eta(x)\eta(y)\in E(G)$ \\
			            and $\alpha\big(\eta(x)\big)\alpha\big(\eta(y)\big) \in E(H)$
			      \item[(ii)] $\Big(b_G\big(\eta(x)\eta(y)\big), 0\Big)$
			            if $\eta(x)\eta(y)\in E(G)$
			            and  $\alpha\big(\eta(x)\big)\alpha\big(\eta(y)\big) \notin E(H)$
			      \item[(iii)] $\bigg(0, b_H\Big(\alpha\big(\eta(x)\big)\alpha\big(\eta(y)\big)\Big)\bigg)$
			            if $\eta(x)\eta(y) \notin E(G)$ \\
			            and $\alpha\big(\eta(x)\big)\alpha\big(\eta(y)\big) \in E(H)$
		      \end{enumerate}

	\end{enumerate}
\end{definition}

\subsection{Feasibility as ITS Graph Property}
\label{asec:rfasitsporp}

\begin{figure}[t!bh]
	\centering
	\includegraphics[width=\figwidth\textwidth]{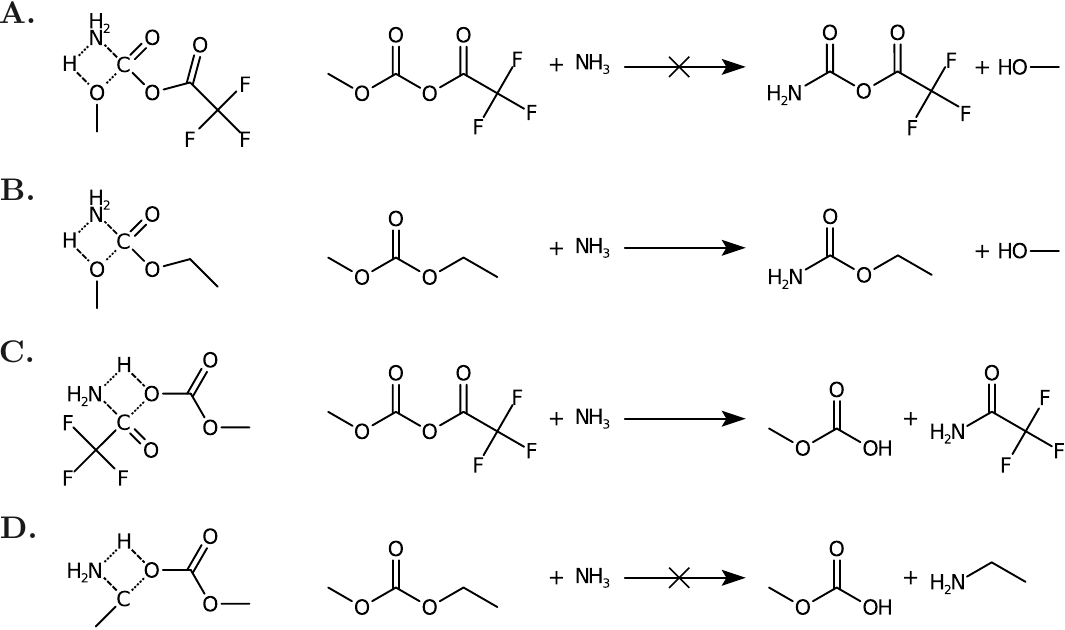}

  \caption{Counterexample for reaction feasibility being a monotonic
  property $\mathcal{P}$ of the ITS graph. The reactions in A and D are
  generally infeasible reactions, and the reactions in B and C are
  generally feasible. Hence, the ITS graphs in A and D have the property
  $\neg\mathcal{P}$ and the ITS graphs in B and C have the property
  $\mathcal{P}$. The ITS graph in B is a subgraph of A and the ITS graph in
  D is a subgraph of C. The example A and B shows that $\neg\mathcal{P}$ is
  a non-monotonic property and C and D shows that $\mathcal{P}$ is a
  non-monotonic property, because the property is not closed under node and
  edge deletion.}

	\label{afig:rf_monotonicity_counterexample}
\end{figure}

Reaction feasibility can be seen as a global property of the ITS
graph. A graph-property $\mathcal{P}$ is said to be \textit{monotonic} if
it is closed under node and edge deletion (e.g.: planarity, acyclicity).
Equivalently, if $G$ does not satisfy $\mathcal{P}$, then any graph that
contains $G$ as a (not necessarily induced) subgraph does not satisfy
$\mathcal{P}$~\cite{Alon2005}.
Figure~\ref{afig:rf_monotonicity_counterexample} shows a counterexample for
reaction feasibility being a monotonic property of the ITS graph.

\subsection{Algorithms}

Algorithm~\ref{alg:extension_DAG} provides a pseudocode description for the
adapted node-induced subgraph enumeration algorithm proposed by Alokshiya
et al.~\cite{Alokshiya2019}. The algorithm begins by initializing the node
set $U$ with the root node $\alpha$, and the candidate set $C$ with the
direct neighbors of $\alpha$. The nodes in $G$ are assigned ascending
integer indices, starting from 0 at the root, thereby defining a total
order on $V(G):u < v$ iff the index of $u$ is less than the index of $v$.
The extension DAG $\mathcal{D}$ is initialized with a single node
representing $\alpha$, and no edges. Two auxiliary data structures, $D$ and
$P$, are used to store the distance of each node to the root, and the
parent node from where the extension is created. These structures help
prune redundant exploration paths and ensure efficient traversal. For a
comprehensive explanation of the underlying subgraph enumeration algorithm,
refer to the work by Alokshiya et al.~\cite{Alokshiya2019}. The function
\texttt{EnumerateCIS} recursively extends the root node set and constructs
the full extension DAG. The algorithm can be turned into a linear-delay
generator by yielding the newly added node after
line~\ref{lst:line:extension_DAG_yield_line} in
Algorithm~\ref{alg:extension_DAG}.

\begin{algorithm}
	\caption{Algorithm for constructing all rooted connected node-induced
		subgraphs for an input graph $G$ structured as an extension DAG. This
		algorithm builds on the algorithm by Alokshiya et al.~\cite{Alokshiya2019}}
	\label{alg:extension_DAG}
	\begin{algorithmic}[1]
		\Require An undirected graph $G$ and an anchor node $\alpha\in V(G)$
		\OUTPUT Complete extension DAG $\mathcal{D}$

		\State $U := [\alpha]$ and $C := neighbors(\alpha)$
		\State $\mathcal{D} := (\{\{\alpha\}\}, \emptyset )$
		\State $D[\alpha] := 0$ and set $D[c] := 1, P[c] := \alpha \quad
			\forall c \in C$
		\State \Call{EnumerateCIS}{$\mathcal{D}, U, C, D, P$}
		\State \Return $\mathcal{D}$

		\\
		\Function{enumerateCIS}{$\mathcal{D}, U, C, D, P$}
		\For{$v$ in $C\setminus set(U)$} \Comment{$set(U):$ list $U$ treated as set}
		\If{\Call{isValidExtension}{$U, v, D$}}
		\State $C' \gets \{u \in neighbors(v) | u \notin C \cup set(U)\}$
		\State $D' \gets D$ and $P' \gets P$
		\For{$u$ in $C'$}
		\State $D'[u] := D[v] + 1$ and $P'[u] := v$
		\EndFor
		\State $U' \gets U$ appended by $v$
		\State add node $set(U')$ to $\mathcal{D}$
		\State add edge $(set(U'), set(U))$ to $\mathcal{D}$ \label{lst:line:extension_DAG_yield_line}
		\State \Call{enumerateCIS}{$\mathcal{D}, U', C \cup C', D', P'$}
		\ElsIf{\Call{isExistingExtension}{$U, v, D$}}
		\State add edge $(set(U) \cup \{v\}, set(U))$ to $\mathcal{D}$
		\EndIf
		\EndFor
		\EndFunction

		\\
		\Function{isValidExtension}{$U, v, D$}
		\State $s = U[0]$ \Comment{root}
		\State $x = U[-1]$ \Comment{last added}
		\If{$v < s$}
		\Return False
		\ElsIf{$D[v] > D[x]$}
		\Return True
		\Else
		\State\Return $\neg$ \Call{isExistingExtension}{$U, v, D$}
		\EndIf
		\EndFunction

		\\
		\Function{isExistingExtension}{$U, v, D$}
		\State $x = U[-1]$ \Comment{last added}
		\State \Return $D[v] \neq D[x]$ or $v \leq x$
		\EndFunction
	\end{algorithmic}
\end{algorithm}

\subsection{Complexity Analysis}
\label{asec:complexity}

\begin{figure}
	\centering
	\includegraphics[width=\figwidth\textwidth]{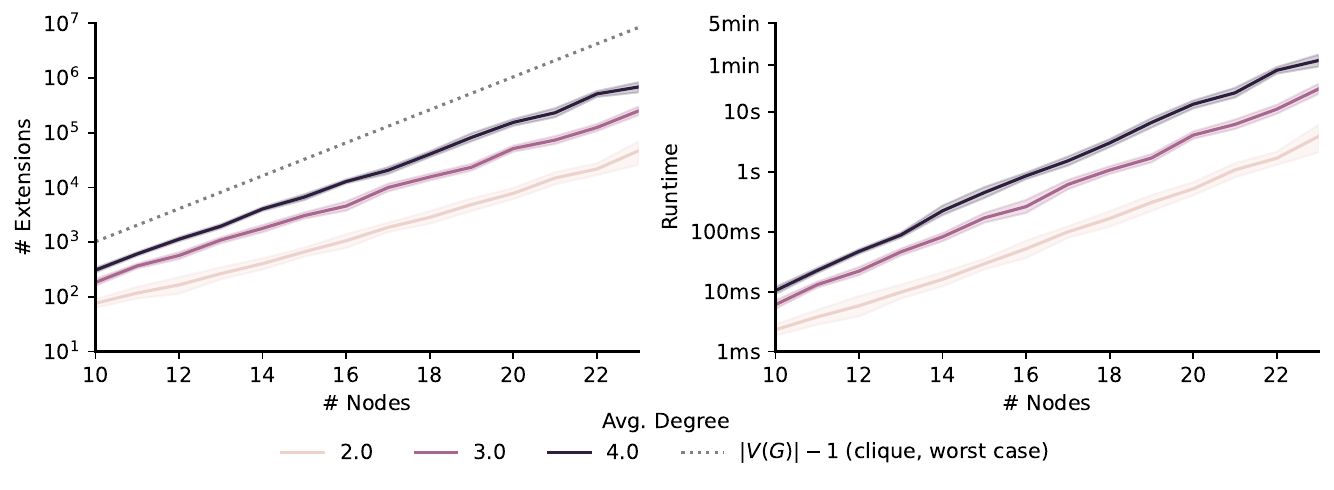}

	\caption{The left plot depicts the number of extensions over the number
		of nodes, conditioned on the average node degree. The dotted line
		indicates the worst case complexity of $2^{|V(G)|}$ possible extensions. The
		right plot shows the mean running time over the number of nodes
		conditioned on the mean average node degree. Running time measurements
		were performed on a Ryzen 3700X CPU. }

	\label{fig:its_extension_complexity}
\end{figure}

ITS graphs typically have low degree and are bounded by a constant upper
degree. Despite this structural simplicity and the imposed constraints, the
computational complexity of finding PI explanations appears to be
exponential in practice. Fig.~\ref{fig:its_extension_complexity} presents
an empirical analysis, showing the number of generated extensions and the
corresponding computation time over the number of nodes in the graph.
The number of subgraphs that need to be considered is bounded above by
$2^{|E(G)|-r}$, where $r$ is the number of edges in the reaction
center. Even though the ITS graph $G$ has bounded degree because the
constituting molecular graph have bounded degree, the number of connected
subgraphs still grows exponentially with $V(G)$ in general
\cite{Kangas:18}.
The observed growth
in computation time and number of extensions suggests that computing
solutions beyond small instances becomes computationally infeasible with
the current approach. The small instances, however, include some of the
chemically relevant problems.

\end{document}